\def\year{2022}\relax
\documentclass[letterpaper]{article} % DO NOT CHANGE THIS
\usepackage{aaai22}  % DO NOT CHANGE THIS
\usepackage{times}  % DO NOT CHANGE THIS
\usepackage{helvet}  % DO NOT CHANGE THIS
\usepackage{courier}  % DO NOT CHANGE THIS
\usepackage[hyphens]{url}  % DO NOT CHANGE THIS
\usepackage{graphicx} % DO NOT CHANGE THIS
\usepackage{natbib}  % DO NOT CHANGE THIS AND DO NOT ADD ANY OPTIONS TO IT
\usepackage{caption} % DO NOT CHANGE THIS AND DO NOT ADD ANY OPTIONS TO IT
\DeclareCaptionStyle{ruled}{labelfont=normalfont,labelsep=colon,strut=off} % DO NOT CHANGE THIS
\usepackage{algorithm}
\usepackage{algorithmic}

\usepackage{newfloat}
\usepackage{listings}

\usepackage{graphicx}
\usepackage{amsmath}
\usepackage{amsthm}
\usepackage{amssymb}
\usepackage{multirow}
\usepackage{caption}
\usepackage{subcaption}
\usepackage{adjustbox}
\usepackage{mathrsfs}
\usepackage{xcolor}

\newcommand{\norm}[2]{\left\lVert#1\right\rVert_{#2}}
\usepackage{microtype}
\usepackage{mdframed}
\mdfdefinestyle{MyFrame}{%
    linecolor=gray,
    innertopmargin=2pt,%\baselineskip,
    innerbottommargin=2pt,%\baselineskip,
    innerrightmargin=2pt,
    roundcorner=10pt,
    innerleftmargin=2pt,
    linewidth=1pt,
    backgroundcolor=white
    }
    
\newtheorem{definition}{Definition}
\newtheorem{theorem}[definition]{Theorem}

\definecolor{codegreen}{rgb}{0,0.6,0}
\definecolor{codegray}{rgb}{0.5,0.5,0.5}
\definecolor{codepurple}{rgb}{0.58,0,0.82}
\definecolor{backcolour}{rgb}{0.95,0.95,0.92}
\definecolor{lightgray}{gray}{0.9}   

\definecolor{yellow}{RGB}{255,255,153}
\definecolor{grey}{RGB}{224,224,224}

\newboolean{showcomments}
\setboolean{showcomments}{true}
\ifthenelse{\boolean{showcomments}}
 { \newcommand{\mynote}[2]{
      \fbox{\bfseries\sffamily\scriptsize#1}
        {\small$\blacktriangleright$\textsf{\emph{#2}}$\blacktriangleleft$}}}
        { \newcommand{\mynote}[2]{}}

\definecolor{DarkOrange}{rgb}{0.8,0.3,0.0} 
\definecolor{DarkCyan}{rgb}{0.0, 0.55, 0.55}
\definecolor{Gray}{gray}{0.9}

\floatstyle{ruled}
\newfloat{listing}{tb}{lst}{}
\floatname{listing}{Listing}
\title{Adversarial Robustness in Multi-Task Learning: Promises and Illusions}
\author{
    Salah Ghamizi \textsuperscript{\rm 1},
    Maxime Cordy \textsuperscript{\rm 1},
    Mike Papadakis \textsuperscript{\rm 1},
    Yves Le Traon \textsuperscript{\rm 1}
}
\affiliations{
    \textsuperscript{\rm 1}University of Luxembourg\\
    salah.ghamizi@uni.lu, maxime.cordy@uni.lu, michail.papadakis@uni.lu, yves.letraon@uni.lu
}

\usepackage{bibentry}
\begin{document}

\maketitle

%%%%%%%%% ABSTRACT
\begin{abstract}
   Vulnerability to adversarial attacks is a well-known weakness of Deep Neural networks. While most of the studies focus on single-task neural networks with computer vision datasets, very little research has considered complex multi-task models that are common in real applications. In this paper, we evaluate the design choices that impact the robustness of multi-task deep learning networks. We provide evidence that blindly adding auxiliary tasks, or weighing the tasks provides a false sense of robustness. Thereby, we tone down the claim made by previous research and study the different factors which may affect robustness. In particular, we show that the choice of the task to incorporate in the loss function are important factors that can be leveraged to yield more robust models.
   
   We provide all our algorithms, models, and open source-code at \emph{[Anonymized for double-blind review]} 
   
   %, first theoretically, then through the empirical evaluation of five architectures and eleven different tasks. Our results suggest that adding auxiliary tasks provides a false sense of robustness and that the choice of the tasks we incorporate and in which way has a deeper impact on the robustness of the models. 
   %In particular, we show that dynamically adjusting the weights suffice to increase the robustness of the multi-task models.
   %In particular, we show that dynamically adjusting the weights of the loss of individual tasks at each step of the attacks suffice to break the apparent robustness of multi-task learning and causes 39\% more error than the state-of-the-art attack with 30\% less computation.   
\end{abstract}

%%%%%%%%% BODY TEXT
\section{Introduction}

While most research on computer vision focuses on single-task learning, 
%in a mono-task mono-modal context, 
many real applications (especially in robotics \cite{radwan2018vlocnet}, autonomous vehicle \cite{yu2020bdd100k}, privacy
\cite{ghamizi2019adversarial} and medical diagnosis \cite{xu2011survey}) require learning to perform different tasks from the same inputs. %the estimation on many parameters from one input. 
For instance, an autonomous vehicle processes multiple computer vision tasks to navigate properly \cite{midLevelReps2018, savva2019habitat}, such as object segmentation and depth estimation.

To meet such requirements, one can build and train on a specific model for each task. However, previous studies have shown that multi-task learning, i.e. building models that learn to perform multiple tasks simultaneously, yields better performance than learning the individual tasks separately \cite{zhang2017survey,vandenhende2021multi}. The intuition behind these results is that tasks that share similar objectives benefit from the information learned by each other. However, while the performance of multi-task learning approaches on clean images has seen major improvements recently\cite{standley2020tasks}, the security of these models, in particular their vulnerability to adversarial attacks, has been barely studied.

The phenomena of \emph{adversarial attacks} has first been introduced by \cite{Biggio2013} and \cite{Szegedy2013} and has since gathered the interest of researchers to propose new attacks \cite{fgsm_goodfellow2015explaining, pgd_madry2019deep}, defense mechanisms \cite{kurakin2016adversarial, he2017adversarial}, detection mechanisms \cite{metzen2017detecting}, or to improve transferability across different networks \cite{tramer2017space, inkawhich2019feature}.  

~\cite{mao2020multitask} pioneered the research on adversarial attacks against multi-task models. Their main result is that, under specific conditions, making models learn additional \emph{auxiliary} tasks leads to both an increase in clean performances \emph{and} improves the robustness of models.

In this paper, we pursue the endeavor of Mao et al. and study whether their conclusions hold in a larger spectrum of settings. Surprisingly, our experimental study shows that adding more tasks does not consistently increase robustness, and may even have negative effects. We even reveal some experimental parameters, such as the attack norm distance, can annihilate the validity of the previous theoretical results. Overall, this indicates that increasing the number of tasks only gives \emph{a false sense of robustness}.

In face of this disillusionment, we investigate the different factors that may explain the discrepancies between our results and that of the previous study. We demonstrate that the contribution of each task to the vulnerability of the model can be qualified, and that what matters is not the number of added tasks but the \emph{marginal vulnerability} of these tasks.

Following this finding, we investigate remedies to reinstate the addition of auxiliary tasks as an effective means of improving robustness. Firstly, following the recommendation of previous research on increasing the clean performance of multi-task models \cite{chen2018gradnorm, standley2020tasks}, we show that adjusting the task weights to make vulnerable tasks less dominant yields drastic robustness improvement but does so only against non-adaptive adversarial attacks: Auto-PGD \cite{autoattack} and Weighted Gradient Descent -- a new adaptive attack that we propose to adapt the produced perturbation to task weights -- annihilate the benefits of weight adjustment. 

Secondly, we show that a careful selection of the tasks suffices to ensure that model robustness increases. Given a target model, determining which combination of tasks is optimal can be costly. We propose different methods to approximate the gain in robustness and show that they strongly correlate with the robustness of the target model.%real robustness increase (after adding the suggested tasks).

%Following this finding, we investigate if optimal task weighing can improve the robustness of multi-task models. \cite{chen2018gradnorm, standley2020tasks} has shown that tasks weight optimization is an effective means of improving clean performance. Our research confirms that weight optimization effectively improve the robustness against non-adaptive attacks. Weight optimization however does resist to adaptive attacks and we show that adaptive attacks like Auto-PGD \cite{autoattack} as well as WGD, our proposed adaptive attack cause significant errors both on weighted and non-weighted multi-task models and that only some combinations of tasks remain robust to adaptive attacks.

%Finally, given a target model, determining which combination of tasks is optimal is costly. We propose various methods to approximate the gain in robustness and show that they strongly correlate with the robustness of the target model.

To summarize, our contributions are:
\begin{itemize}
    \item We replicate the study of Mao et al. under a broader set of experimental settings.
    \item We show that adding auxiliary tasks is not a guarantee of improving robustness and identify the key factors explaining the discrepancies with the original study.
    \item We refine the theory of Mao et al. through the concept of \emph{marginal adversarial vulnerability} of tasks. Leaning on this, we demonstrate that the inherent vulnerability of tasks plays a central role in the model robustness.
    \item We empirically show that a careful selection of the tasks can act as a remedy and offer the benefits initially promised by previous research. However, weighting the tasks does not provide increased robustness against adaptive attacks.
    \item We propose a set of surrogates to efficiently evaluate the robustness of a combination of tasks.
\end{itemize}

\section{Related Work}

\paragraph{Multi-task learning (MTL).}

Multi-task learning leverages shared knowledge across multiple-tasks to learn models with higher efficiency \cite{vandenhende2021multi,standley2020tasks}. 
A multi-task model is commonly made of an encoder block, that learns shared parameters across the tasks and a decoder part that branches out into %multiple 
task-specific heads. 

%To answer the first, question, MSTL networks historically belonged either to the "Hard Parameter sharing" family (UberNet[], Stochastic Filter Grouping []) where the parameter sharing is only restricted to the encoder and each head of the decoder learns freely its own parameters, or "Soft Parameter sharing" family where the heads still have some level of interactions and constraints (linear combination of the activations for Cross-stitch Networks [], feature fusion layers for NDDR-CNN [] or attention mechanisms for MTAN [], ...).

\cite{vandenhende2021multi} recently proposed a new taxonomy of MTL approaches to split the approaches based on where the task interactions take place. They differentiated between approaches that are \emph{encoder-focused} where some information is shared across tasks at the encoder stage and approaches that are \emph{decoder-focused} where some interactions still happen across the heads of the tasks. 
%They compared different optimization techniques proposed to improve the learning. 
They organized MTL research around three main questions: (1) when does the task learning interact, (2) how can we optimize the learning, and (3) which tasks should be learned together. 

%To answer (2), Vandenhende surveys the different optimization techniques from the literature and point out that most of the optimizations focus on the task balancing. In the MTSL context, we need to take into account that the loss of individual tasks may be imbalanced, which may cause some tasks to have a dominant impact in the joint learning process and hence harm the learning of the other tasks. Some of the most common strategies are to choose ad-hoc weights based on fine-tuning, use Gradient normalization [], Uncertainty weighing [] or Multi-objective optimizations. (Gong et al., 2019) and (Leang et al.,2020) both show there is no clear winner and similar performance among strategies,including a uniform weighting strategy.

%While (1) and (2) have an extensive literature with well established approaches, answering (3) remains challenging. Prior work evaluated the similarity of tasks based on how does the learning transfer from one to another [Taskonomy, (Pal \& Balasubramanian, 2019; Dwivedi \&Roig., 2019; Achille et al., 2019; Wang et al., 2019], a more recent work showed that learning transfer affinity and multi-task learning affinity display notable differences [Which tasks should be learned together]. 

Our work complement this research by tackling a fourth question: How to optimize the robustness of MTL. 

%focuses on design approaches from the "encoder-focused/ Hard sharing" family, we will evaluate different optimization techniques (no weighing, fixed weights, gradient normalization weights), and we introduce the new concept of \emph{multi-task robustness affinity} and we evaluate its link with some of the previously proposed task affinities. 

\paragraph{Adversarial attacks}

An adversarial attack is the process of intentionally introducing perturbations on the inputs of a machine learning model to cause wrong predictions. One family of adversarial attacks is \emph{poisoning attacks}~\cite{biggio2012poisoning} where the inputs targeted are the training set and occur during the learning step, while \emph{evasion attacks} \cite{Biggio2013} focus on the inference step. 

One of the earliest attacks is the Fast Gradient Sign Method (FGSM) \cite{fgsm_goodfellow2015explaining}. It adds a small perturbation \(\eta\) to the input of a neural network, which is defined as:
\begin{equation}
\eta = \epsilon \, \text{sign}(\nabla_x \mathscr{L}_i(\theta,x,y_i)),\label{eq:fgsm}
\end{equation}
where \(\theta\) are the parameters of the network, \(x\) is the input data, \(y_i\) is its associated target, \(\mathscr{L_i}(\theta,x,y_i)\) is the loss function used, and \(\epsilon\) the strength of the attack. %Note that the required gradient is easily computed through backpropagation if the network structure is known (white-box).
%FGSM is a single step attack. 
Following Goodfellow, other attacks were proposed, first by adding iterations (I-FGSM)~\cite{kurakin2016adversarial}, projections and random restart (PGD)~\cite{pgd_madry2019deep}, momentum (MIM)~\cite{dong2018boosting} and constraints (CoEva2) ~\cite{coeva2}.

These algorithms can be used without any change on a multi-task model if the attacker only focuses on a single task.

%Recent work on adversarial attacks tackled dense predictions models and multi-task models.\cite{arnab2018robustness} suggested that single-step attacks suffice to attack weak datasets like Cityscapes\footnote{in the sense with less variability and similar tasks} and that network components (residual connections, ...) contribute to the robustness of models. Our work diff as we evaluate models with multiple unrelated tasks and evaluate other design choices. 

\cite{mao2020multitask} were first to evaluate these gradient-based attacks. They extended the first-order vulnerability found by \cite{simon2019first} and showed that increasing the number of tasks leads to more robust models.

%We argue that while the first-order expansion describes well the adversarial vulnerability under $l_2$ attacks, it does not suffice to explain the weakness of multi-task models under $l_\infty$ multi-step attacks.
%We show that this finding does not generalize both to L-inf and L2-distance attack with limited steps, show that we need to go beyond the first order-approximation to explain the apparent robustness of the models and propose an optimized attack that counter this apparent robustness.

%Finally, we focus our research not only on how many tasks we should add, but how to choose these tasks and whether existing taxonomies (based on learning or transfer) help in choosing the robust tasks.
%According to Zhang [An overview of multi-task learning], 3 challgens need to be solved

%------------------------------------------------------------------------

\section{Problem formulation}

\subsection{Preliminaries}

Let $\mathscr{M}$ a multi-task model with tasks $\mathscr{T}= \{t_1,...,t_M\}$. For each input example $x$, we denote by $\bar{y}$ the corresponding ground-truth label and we have $\bar{y} = (y_1, ..., y_i, y_M)$ where $y_i$ is the corresponding ground truth for task $i$. 

%In a multi-task context, for each $x$, $y_i$ is the corresponding ground truth for task $i$ and we have $\bar{y} = (y_1, ..., y_M)$.

We focus on hard sharing multi-task learning, where the models are made of one encoder (backbone network common to all tasks) $E(.)$ and task-specific decoders $D_i(.)$. Each task is associated with a loss $\mathscr{L}_i$; $\mathscr{L}_i(x,y_{i}) = l_i(D_i(E(x)),y_i)$
where $l_i$ is a loss function tailored to the task $i$. For instance, we can use cross-entropy losses for classification tasks, and L1 losses for regression tasks.

The total loss $\mathscr{L}$ of our multi-task model is a weighted sum of the individual losses $\mathscr{L}_i$ of each task:
$$ \mathscr{L}(x,\bar{y}) = \sum_{i=1}^{M} w_i \mathscr{L}_i(x, y_i)$$
where $\{w_1 \dots w_M\}$ are the weights of the tasks, either set manually or optimized during training \cite{chen2018gradnorm}.

\paragraph{Single-task adversarial attacks}

In this use case, the adversary tries to increase the error of one single task. This threat model can represent scenarios where the attacker has access to one task only or aims to perturb one identified task.  
The objective of the attacker can then be modeled as:

\begin{equation}
\underset{\delta}{\operatorname{argmax}} \: \mathscr{L}_i(x+\delta,y_i) \: \text{s.t.} \: \norm{\delta}{p} \leq \epsilon
\label{eq:single-task-objective}
\end{equation}
where $i$ is the index of the targeted task and $\epsilon$ the maximum perturbation size using a norm $p$.

\paragraph{Multi-task adversarial attacks}

In multi-task adversarial attacks, the adversary aims to increase the error of multiple outputs all at once. This captures scenarios where the adversary does not have fine-grained access to the individual tasks or where the final prediction of the system results from the combination of multiple tasks. Therefore, the adversary has to attack all tasks together. 

Given a multi-task model $\mathscr{M}$, an input example $x$ , and $\bar{y}$ the corresponding ground-truth label, 
the attacker seeks the perturbation $\delta$ that will maximize the joint loss function of the attacked tasks -- i.e. the summed loss, within a p-norm bounded distance $\epsilon$. The objective of the attack is then:

\begin{equation}
 \underset{\delta}{\operatorname{argmax}} \: \mathscr{L}(x+\delta,\bar{y}) \: \text{s.t.} \: \norm{\delta}{p} \leq \epsilon 
\label{eq:multi-task-objective}
\end{equation}

%While the interactions between different tasks are negligible when a single-task is targeted (as shown in section XXX), attacking multiple tasks together brings the interactions between tasks into play and, thereby, affects the success rate of the attack.

\paragraph{Adversarial vulnerability of multi-task models}

%We provide the proofs for our lemmas and theorems in the Appendix A.

\cite{simon2019first} introduced the concept of adversarial vulnerability to evaluate and compare the robustness of single-task models and settings. Mao et al. extended it to multi-task models as follow:
%\cite{yang2020multitask} applied the same definition and approximation to multi-task models. We extend the adversarial vulnerability to the second order and show later empirically that there is no one-to-one relationship between the average norms of the gradients and adversarial vulnerability in multi-task models and hence, the need to go beyond the first order approximation.  

%In this section, we provide a theoretical evaluation of the adversarial vulnerability of multi-task models. The proofs are provided in the supplementary materials.

\begin{definition}

%\paragraph{Definition 1.} 
Let $\mathscr{M}$ be a multi-task model. $\mathscr{T}' \subseteq \mathscr{T}$ a subset of its tasks and $\mathcal{L}_{\mathscr{T}'}$ the joint loss of tasks in $\mathscr{T}'$. Then, we denote by $\mathbb{E}_{x}[\delta\mathcal{L}(\mathscr{T}',\epsilon)]$ the \emph{adversarial vulnerability} of $\mathscr{M}$ on $\mathscr{T}'$ to an $\epsilon$-sized $\| .\|_p$-attack, and define it as the average increase of $\mathcal{L}_{\mathscr{T}'}$ after attack over the whole dataset:%, i.e.:
%with $\mathscr{T}= \{t_1,...,t_M\}$ tasks, an input $x$, $\bar{y} = (y_1, ..., y_M)$ its corresponding ground-truth.
%We denote the set of attacked tasks $\mathscr{T}'$, a subset of the model's tasks $\mathscr{T}$. and the joint task loss of attacked tasks $\mathcal{L}'$. Then, we define the \emph{adversarial vulnerability} of $\mathscr{M}$ on $\mathscr{T}' \subseteq \mathscr{T}$ to an $\epsilon$-sized $\| .\|_p$-attack, the average increase-after attack over the whole dataset of the joint loss of the attacked tasks, i.e.:

%\scalebox{0.7}{

$$    \mathbb{E}_{x}[\delta \mathcal{L}(\mathscr{T}',\epsilon)]= \\ \mathbb{E}_{x}\left[\max _{\|\delta\|_{p}\leq\epsilon} \mid \mathcal{L}_{\mathscr{T}'}(x+\delta, \bar{y})-\mathcal{L}_{\mathscr{T}'}(x, \bar{y}) \mid \right]
$$
\end{definition}

This definition matches the definitions of previous work \cite{goodfellow2015explaining, sinha2017certifying} of the robustness of deep learning models: the models are considered vulnerable when a small perturbation causes a large average variation of the joint loss.

Similarly, we call \emph{adversarial task vulnerability} of a task $i$ the average increase of $\mathcal{L}_{\mathscr{T}'}(x,y_i)$ after an attack over the whole dataset.

%Assuming that the variation $\delta$ is small, \cite{simon2019first} proposed the following Taylor expansion in $\epsilon$:

% préciser clairement les assumptions: surtout que la first taylor est une assumption

%\begin{lemma} 
%Under an $\epsilon$-sized $\| .\|_p$-attack, the adversarial vulnerability of a multi-task model can be approximated through the first-order Taylor expansion, that is:
%\scalebox{0.7}{
%\begin{equation}
%    \mathbb{E}_{x}[\delta \mathcal{L}'(x, \bar{y}, \epsilon, \mathscr{T}')]\approx \epsilon \cdot \mathbb{E}_{x}[\left\mid\mid \partial_{x} \mathcal{L}'(x, \bar{y}) \mid \mid_q\right] 
    %+ \\ \frac{\epsilon^2}{2} \cdot \mathbb{E}_{x}[\left\mid\mid \partial^2_{x} \mathcal{L}'(x, \bar{y}) \mid \mid_q\right]
%\end{equation}
%\label{lemme2}
%\end{lemma}
%\begin{proof}
%\vskip -0.3in
%Appendix A.1
%\end{proof}

%They showed that this first order expansion is a sufficient approximation of the adversarial vulnerability for the single tasks they evaluated.

Assuming that the variation $\delta$ is small, \cite{mao2020multitask} proposed the following theorem using the \cite{simon2019first} first-order Taylor expansion in $\epsilon$:

\begin{theorem}
\label{thm:mao}

Consider a multi-task model $\mathscr{M}$ where an attacker targets $\mathscr{T}= \{t_1,...,t_M\}$ tasks uniformly weighted, with an $\epsilon$-sized $\| .\|_p$-attack. If the model is converged, and the gradient for each task is i.i.d. with zero mean and the tasks are correlated, the adversarial vulnerability of the model can be approximated as

\begin{equation}
\begin{aligned}
    \mathbb{E}_{x}[\delta \mathcal{L}']\approx K \cdot \sqrt{\frac{1+\frac{2}{M} \sum_{i=1}^{M} \sum_{j=1}^{i-1} \frac{\operatorname{Cov}\left(\mathbf{r}_{i}, \mathbf{r}_{j}\right)}{\operatorname{Cov}\left(\mathbf{r}_{i}, \mathbf{r}_{i}\right)}}{M}}
\end{aligned}
\end{equation}
\label{theorem3}
\end{theorem}
where K is a constant dependant of $\epsilon$ and the attacked tasks, and $\mathbf{r}_{i}=\partial_{x} \mathcal{L}(x, y_i)$ the gradient of the task i, and $\operatorname{Cov}\left(\mathbf{r}_{i}, \mathbf{r}_{j}\right)$ the covariance between the two gradients $\mathbf{r}_{i}, \mathbf{r}_{j}$.

\begin{proof}
Appendix A.1 and A.2
\end{proof}

This theorem indicates that under the specific assumptions described above, (1) increasing the number of tasks reduces the adversarial vulnerability of a multi-task model and (2) even more when these tasks are uncorrelated.

\subsection{Research Questions and Methodology}

Our research endeavor stems from the hypothesis that the assumptions supporting the above theorem are too restrictive to be met in practice. The existence of settings where these assumptions do not hold would tone down the validity of the results of \cite{mao2020multitask} and raise anew the question of how to achieve robust multi-task learning.

Accordingly, our first research question investigates if the assumptions and results of Theorem \ref{thm:mao} are confirmed in a variety of multi-task models and settings. We ask:

\begin{description}
\item[\textbf{RQ1:}] \emph{Are multi-task models reliably more robust than single-task models?}
\end{description}
To answer this question, we study whether the results of Theorem \ref{thm:mao} (i.e. adding tasks increases robustness) generalize to other settings. In particular, 
%Yang et al. \cite{yang2020multitask} were 
Theorem \ref{thm:mao} was proven for $l_2$-normed attacks and we investigate if their results remain valid for $l_\infty$-normed attacks. Finally, we investigate the impact of various experimental settings covering different perturbation budgets $\epsilon$, architectures, and number of training steps.

Following this, we formulate an alternate hypothesis that could explain the 
%discrepancies between the promises of Yang et al. \cite{yang2020multitask}
apparent robustness of multi-task models
and the evidence brought by our study: what matters most is not the number of tasks or how they correlate but how much the tasks individually impact the vulnerability of the model. Thus, making more robust a model with one task that is ``marginally more vulnerable'' requires adding robust tasks that make this model ``marginally less vulnerable''.

Our second research question studies this hypothesis and attempts to quantify this \emph{marginal vulnerability} :

\begin{description}
\item[\textbf{RQ2:}] \emph{How to quantify the individual contribution of each task on the robustness of the model?}
\end{description}
To answer this question, we define the concept of marginal adversarial vulnerability of a model to a task $i$ as the variation between the adversarial vulnerability of the model with this newly added task $i$ and its vulnerability before this task.

We, then, formally provide an upper bound of how much a single-task impacts the vulnerability of a multi-task model and empirically show that specific tasks can reduce the robustness of the model whatever the number of tasks.

%Understanding how individual tasks can deteriorate the model's robustness raises the following question:

%Our theoretical and empirical analyses show that some tasks help improving the robustness of the model, but that generally adding more tasks does not suffice to balance some very weak tasks. Multi-task learning literature suggests that the correct weighing of tasks is of utmost importance for optimal performances \cite{vandenhende2021multi}. We hypothesise a parallel thinking about robustness and claim that one can significantly improve model robustness by adjusting the weights of the tasks or by carefully selecting the tasks to add. We investigate the first option and ask: 

Based on this concept of marginal adversarial vulnerability, we look for ways to %further 
improve robustness. Past research on multi-task learning suggests that carefully weighing tasks can drastically improve clean performance \cite{vandenhende2021multi}. Leaning on this idea, we hypothesize that one can improve model robustness by adjusting the weights of the tasks. %or by carefully selecting the tasks to add. 
We ask:

\begin{description}
\item[\textbf{RQ3:}] \emph{Can one improve the robustness of multi-tasks models by adjusting the weights of the tasks?}
\end{description}

We show that optimizing the weights significantly improves the robustness of multi-tasks models against PGD. However, this apparent robustness may actually result from a gradient masking effect \cite{athalye2018obfuscated} caused by the weight adjustment. We, therefore, investigate if adaptive attacks -- which are known to circumvent gradient masking -- can successfully attack the weight-optimized model. We use Auto-PGD \cite{autoattack} and propose a new attack that adjusts at each step of the attack which tasks are attacked and how much the gradient of each task is penalized to compute the optimal perturbation.

Finally, we investigate the practical question of how to identify the combinations of tasks that yield the highest robustness. Indeed, training multi-task models can be expensive and our second research question shows that choosing the right combination of tasks is critical. Given two multi-task models, we propose a set of guidelines that help practitioners to infer the robustness of the task combinations of each model
%of expensive combinations 
from cheaper models. Our final question is:  

\begin{description}
\item[\textbf{RQ4:}] \emph{How to efficiently find combinations of tasks giving the best robustness?}
\end{description}

\subsection{Experimental setup}

%\begin{figure}[]
%\begin{center}
%\centerline{\includegraphics[width=\columnwidth]{Figures/task_tree.jpg}}
%\caption{Task similarity tree proposed by Zamir et al.\cite{taskonomy2018}. D and d tasks are adjacent, E is farther in the tree leaves, then n and s is the farthest leave from d. 
%Our aim was to cover a wide field of tasks then evaluate if this taxonomy has any impact on the robustness of the combined tasks.
%We focus our figures on these 5 tasks to cover a diverse set of tasks.
%}
%\label{taskonomy-tree}
%\end{center}
%\vskip -0.3in
%\end{figure}

The following describes our general experimental setup used across all RQs. It must be noted that the setups specific to the RQs are presented in their dedicated sections.

\textbf{Dataset.} We use the Taskonomy dataset, an established dataset for multi-task learning \cite{taskonomy2018}. From the original paper, we focus on 11 tasks : Semantic Segmentation (s), Depth z-buffer Estimation (d), Depth euclidian Estimation (D), Surface Normal Prediction (n), SURF Keypoint Detection in 2D (k) and 3D (K), Canny Edge Detection (e), Edge Occlusion (E), Principal Curvature (p), Reshading (r) and Auto-Encoders (A). This subset of tasks is at the intersection of the major studies \cite{taskonomy2018, vandenhende2019branched, standley2020tasks} about tasks similarity. %We restrict our study to the "small" split of Taskonomy to match the experimental protocol of \cite{mao2020multitask}. 

We chose this dataset because (1) it offers a diverse set of computer vision tasks; (2) it is regularly used in the literature to evaluate task performances and (3) it comes with a taxonomy of tasks where similar tasks are grouped, which is useful for our experiments. Besides, previous work has shown that segmentation-focused datasets like Cityscape\cite{Cordts2016Cityscapes} have less variability and are weak even against single-task attacks\cite{arnab2018robustness}. Hence, these datasets are not a suitable benchmark for our study.    

\textbf{Attacks.}
 We focus our research on gradient-based attacks. In particular, we use as base setting the $l_\infty$ Projected Gradient Descent attack (PGD) \cite{pgd_madry2019deep} with 25 steps attacks, a strength of $\epsilon=8/255$ and a step size $\alpha=2/255$. 
 
 We also study in RQ3 the impact of adaptive attacks on the robustness of multi-task models. We evaluate the robustness of weighted multi-task models against Auto-PGD~\cite{autoattack}, the strongest parameter-free gradient attack. 
 
 Finally, we design a new attack, \emph{WGD}, that takes into account individual task weights and show that multi-task learning is vulnerable against our adaptive attack. 

%We also evaluate other configurations of attacks in dedicated sub-sections:

%\begin{itemize}
%    \item In section \ref{epsilon_evaluation}, we evaluate different strength of PGD attacks with $\epsilon \in \{2/255,4/255,8/255,16/255\}$
%    \item In section \ref{steps_evaluation}, we evaluate different number of steps of the attack with steps $\in [1,50]$. % including the 1 step without random start that covers the FGSM attack.
    %\item We compare the PGD attack with our two optimizations \textbf{MGD} and \textbf{MAGD}.
%\end{itemize}

\textbf{Models.}
%We use a standard encoder-decoder architecture. 
We use the architectures and training settings of the original Taskonomy paper \cite{taskonomy2018}: A Resnet18 encoder and a custom decoder for each task.%, and we extend in section \ref{architecture_evaluation}
%the evaluation to three families of encoders (Xception, Wide-Resnet, and Resnet) and for the latter, evaluate three sizes of encoders (Resnet18, Resnet50 and Resnet152).
 We use a uniform weights, Cross-entropy loss for the semantic segmentation task and an L1 loss for the other tasks. %We train our tasks with uniform weights.
%For each architecture, we train our tasks with uniform weights.%, and in section \ref{weight_evaluation} we extend the study to tasks with weights proposed by \cite{taskonomy2018} for optimal task performances.

%For each architecture, we consider two different multi-task models. In the first one, we trained and evaluated the models on two tasks among the eleven with equal weights. In the second, we weighted the loss of each task according to the common practice\cite{taskonomy2018}. 

%We use as a baseline the single-task models formed from each encoder architectures trained on each of the eleven tasks.

Our evaluation covers all possible combinations of tasks, i.e 935 multi-task models. Each model is a combination of a main task and one or multiple auxiliary tasks. We present in our evaluation the case where the attacker aims to attack only the main task (single-task attacks) and when all the tasks are attacked simultaneously (multi-task attacks).

We provide in the Appendix B. of supplementary material the detailed setup of each setting and the detailed results.% across all of them.% the tasks and settings.

\subsection{Robustness Metrics}
\label{sec:robustness-metrics}

\paragraph{Task robustness}

The common way to evaluate empirically the adversarial robustness of a DNN is to compute the success rate of the attack, i.e. the percentage of inputs for which the attack can produce a successful adversarial example  under a constrained perturbation budget $\epsilon$.
%Reversing the problem, we can also evaluate the minimum $\epsilon$ required to achieve a 100\% success rate, this is empirical robustness. 

%In the following, we extend the success rate to the multi-task setting, and to different type of tasks (not only classification).

%In classification tasks, the success rate $s$ of the attack under a constraint budget $\epsilon$ is defined as:

%\begin{align*}
%s= \frac{\vert \mathscr{D}_a\vert}{\vert \mathscr{D}_c\vert} ; \; \mathscr{D}_a  \subseteq \mathscr{D}_c \\
%\forall x^{(k)} \in \mathscr{D}_a \: \exists \delta \:  \text{s.t.} \\
%C(x^{(k)}+\delta) \ne y^{(k)} \: \text{and} \: \norm{\delta}{p} \leq \epsilon
%\end{align*}

%Translated into a multi-task context. Let $\mathscr{M}$ a multi-task model with $M$ tasks. We can define the success rate $s_i$ of the attack of task $i$ under a constraint budget $\epsilon$:

%\begin{align*}
%s_i(\mathscr{M})= \frac{\vert \mathscr{D}_a\vert}{\vert \mathscr{D}_c\vert} ; \; \mathscr{D}_a  \subseteq \mathscr{D}_c \\
%\forall x^{(k)} \in \mathscr{D}_a \: \exists \delta \:  \text{s.t.} \\
%f_i(\mathscr{M}(x^{(k)}+\delta),y_i^{(k)})>0 \: \text{and} \: \norm{\delta}{p} \leq \epsilon
%\end{align*}

%where $f_i$ is a task-specific error metric. For classification tasks, we can use a binary comparison as a distance metric to fall-back to the original definition. We will rely in this study on a thresholded MSE for dense and regression tasks.  

While this metric is suited for classification tasks that rely on classification accuracy, most dense tasks rely on metrics where a success rate is hard to define objectively or requires a hand-picked threshold. For instance, ``image segmentation'' uses Intersection Over Union (IoU, between 0 and 1) as a metric, while ``pose estimation'' relies on the number of correct keypoints and their orientation, and ``depth estimation'' uses the mean square error. To account for this diversity of metrics we define a generic metric that reflects how much the performance has degraded (how much relative error) after an attack: \emph{the relative task vulnerability} metric. We use this metric to present our results throughout our experiments, whereas Appendix C presents our results measured with task-specific metrics. 

%One limitation of these metrics is that they require the use of a threshold to define whether the attack is successful or not. This makes it impractical to compare different tasks irrespective of a particular threshold. To overcome this limitation, we define a new metric named the \emph{relative task vulnerability metric}.% and Collaborative task robustness

\paragraph{Relative task vulnerability}

Given a model $\mathscr{M}$, we define the relative task vulnerability $v_i$ of a task $i$ as the average relative error increase when changing clean inputs $x^{(k)}$ into adversarial inputs $x^{(k)}+\delta$, given their associated ground truth $y_i^{(k)}$. Hence, $v_i$ is given by:  
%will be used in our empirical 
%study to evaluate the impact of different design architectures on the robustness of a task in a multi-task model.

\begin{equation*}
v_i = \mathbb{E}_{k}\left[\frac{f_i(x^{(k)}+\delta,y_i^{(k)}) - f_i(x^{(k)},y_i^{(k)})}{f_i(x^{(k)},y_i^{(k)})}\right]
\label{eq:relative-task-robustness}
\end{equation*}

\noindent where error function $f_i(x^{(k)},y_i^{(k)})$ is a task-specific error between the ground truth $y_i^{(k)}$ and the predicted value of $x^{(k)}$ (e.g., MSE, 1-IoU, etc.).

The concept of relative task vulnerability enables the comparison of two models in terms of the robustness they achieve for any of their task(s). %Given a task $i$, comparing two values of $v_i$ allows to compare two settings and evaluate which is more robust. 
A model with a smaller value of $v_i$ indicates that it is more robust to an attack against task $i$.

%In addition to task vulnerability, we provide the results using the task-specific metrics in appendix C.

%\paragraph{Collaborative task robustness}

%While relative task vulnerability helps to evaluate how robust is a specific multi-tasks model, it does not reflect how tasks interact and impact the model's robustness.
%Let $\mathscr{T} = \{t_1,...,t_n\}$ a set of tasks which we want to study and $\mathscr{M}_{i,j}$ a model trained with the two tasks ${t_i, t_j}; i \neq j$

%we evaluate the collaborative robustness of task $t_i$ as:

%$$ c_i = \frac{E_{k,j}[r_j(\mathscr{M}_{j,k})]}{E_k[r_i(\mathscr{M}_{i,k})]} $$

%\input{LaTeX/-attacks}

%------------------------------------------------------------------------
%\input{LaTeX/-experimental_setup}

%------------------------------------------------------------------------
\section{RQ1: Adding Auxiliary Tasks}% VS single-task models}
\label{sec:rq1}

\textbf{Theoreom \ref{thm:mao}} showed that adversarial vulnerability decreases with the number of uncorrelated tasks. We argue that the results do not generalize to different settings (norms, attack strength, ...) and investigate the other factors which may affect the robustness of the models. This allows us to identify the key factors confirming or refuting the results of Theorem3. More precisely, we consider the attack norm $p$, the perturbation budget $\epsilon$, the model architecture, and the number of training steps (the convergence of learning). We summarize our findings below while detailed results are in Appendix C.

\paragraph{Attack norm $p$.} 
\label{norm_evaluation}
We evaluate the relative task vulnerability against single-task and multi-task attacks with a limited perturbation budget ($\epsilon$ = 8/255, 25 attack steps) under $l_2$ and $l_\infty$ attacks. Under $l_2$ attacks, the previous conclusions of \cite{mao2020multitask} are confirmed.
Under $l_\infty$ attacks, however, adding auxiliary tasks does not reliably increase robustness against neither single-task nor multi-task adversarial attacks. We indeed observe for each task $t$ that the single-task model of $t$ is not more vulnerable than multi-task models with $t$ as the main task -- regardless of the fact that a single-task attack or a multi-task attack was used (see, e.g., also Table \ref{table:tasks-opt}). %the average relative task vulnerability of \textbf{n} for instance, increases from 4.5 to 5.9 going from single-task to multi-task, meanwhile 
%The vulnerability of tasks \textbf{s}, \textbf{d} and \textbf{n} overlap between multi-task and single-task attacks under $l_\infty$ attacks (the mean vulnerability of task \textbf{n} even increases in multi-task attacks from 4.5 to 5.9).

\paragraph{Attack budget $\epsilon$.} 
\label{epsilon_evaluation}
We evaluate the robustness of multitask models against attack size $\epsilon \in \{4/255,8/255,16/255,32/255\}$
Under strong adversarial attacks ($\epsilon > 4/255$), multi-task learning does not provide reliable robustness both against single-task and multi-task adversarial attacks. 

\paragraph{Model architecture.} 
\label{architecture_evaluation}
We evaluate the vulnerability of multi-task models on three families of encoders (Xception, Wide-Resnet, and Resnet) and for the latter, three sizes of encoders (Resnet18, Resnet50 and Resnet152). We train each architecture on a pair of tasks from {s, d, D, E, n}. We evaluate the robustness of each combination of tasks (multi-task models) and compare it with the robustness of the same architecture trained using only one of the task (single-task models).  
For all architectures, multi-task models are not reliably less vulnerable than single-task models. On the contrary, when one task is targeted, 80\% of the multi-task models using Resnet50 and Resnet152 architectures are more vulnerable than their single-task counterparts. %30\% for Xception and 25\% for Wide-Resnet     

\begin{mdframed}[style=MyFrame]
\textbf{
Answer to RQ1:}
For large perturbation budgets $\epsilon$, $l_\infty$ norms, or large models, multi-task learning does not reliably improve the robustness against adversarial attacks.
\end{mdframed}
\section{RQ2: Marginal Adversarial Vulnerability}

\subsection{Theoretical analysis}
To better understand the impact of additional tasks on the multi-task vulnerability, we define the concept of marginal adversarial vulnerability of tasks, and propose a new theorem that bounds the contribution of the additional tasks to the model's vulnerability. 

\begin{definition}

Let $\mathscr{M}$  be a multi-task model with $\mathscr{T} = \{t_1,...,t_M\}$ tasks, an input $x$, $\bar{y} = (y_1, ..., y_M)$ its corresponding ground-truth.
We denote the set of attacked tasks $\mathscr{T}_{N}$ and $\mathscr{T}_{N+1}$, two subsets of the model's tasks $\mathscr{T}$ such as $\mathscr{T}_{N+1} = \mathscr{T}_{N} \cup \{t_{N+1}\}$ and $N+1 \leq M$, and let $\mathcal{L}'$ be the joint task loss of attacked tasks.

We define \emph{marginal adversarial vulnerability} of the model to an $\epsilon$-sized $\| .\|_p$-attack as the difference between the adversarial vulnerability over the task set $\mathscr{T}_{N+1}$ and the adversarial vulnerability over the task set $\mathscr{T}_{N}$. It is given by:

\small
\begin{equation*}
    \left.\Delta_{N} \mathbb{E}_{x}[\delta \mathcal{L}'] = \mathbb{E}_{x}[\delta \mathcal{L}'(x, \bar{y}, \epsilon, \mathscr{T}_{N+1})] - \mathbb{E}_{x}[\delta \mathcal{L}'(x, \bar{y}, \epsilon, \mathscr{T}_{N})] \right]
\end{equation*}
\normalsize
\end{definition}

Similarly to the adversarial vulnerability of a model, for a small $\delta$ value we propose to use Taylor expansions to approximate the marginal vulnerability of the model to a given task. We propose the following theory for the marginal change of the vulnerability of a model when we add a task:

\begin{theorem}

For a given multi-task model $\mathscr{M}$, let $\mathbf{r}_{i}=\partial_{x} \mathcal{L}(x, y_i)$ the gradient of the task i, with a weight $w_i$ and zero mean such as the joint gradient of $\mathscr{M}$ is defined as $\partial_{x} \mathcal{L}(x, \bar{y}) = \sum_{i=1}^{M} w_i \mathbf{r}_{i}$. The first-order approximation of the marginal vulnerability is bounded as follow: 

\begin{equation*}
\begin{aligned}
    \widetilde{\Delta_{N} \mathbb{E}_{x}[\delta \mathcal{L}']} \leq \epsilon \cdot ((N+1) \cdot w_{N+1}\mathbb{E}_{x}[\mid \mid \mathbf{r}_{N+1} \mid \mid] +  \\
    N \cdot \max _{i<N+1} w_{i}\mathbb{E}_{x} [\mid \mid \mathbf{r}_{i} \mid \mid] )
\end{aligned}
\label{theorem-marginal-vuln}
\end{equation*}

\end{theorem}

\begin{proof}
Appendix A.4 and A.5
\end{proof}

When all the tasks have the same weight, $w_i = \frac{1}{N}$ and $w_{N+1} = \frac{1}{N+1}$ and we have:

\begin{equation*}
\widetilde{\Delta_{N} \mathbb{E}_{x}[\delta \mathcal{L}']} \leq \epsilon \cdot (\mathbb{E}_{x}[\mid \mid \mathbf{r}_{N+1} \mid \mid] + \max _{i<N+1} \mathbb{E}_{x} [\mid \mid \mathbf{r}_{i} \mid \mid] )
\end{equation*}

%While this theorem relies on the first-order approximation of the Taylor expansion, we can get a more precise bound with higher orders.
This theorem shows that the increase in adversarial vulnerability when we add more tasks does not depend on the number of tasks already attacked but relates to how robust is the new task we are adding and how weak is the most vulnerable task of the model.

%Both theorems 4 and 6 suggest that we should care more about the robustness of auxiliary tasks we are adding to the multi-task model than the number of tasks we add.

\subsection{Empirical evaluation}

To confirm our hypothesis that increasing the number of tasks does not guarantee the increase of robustness, we empirically evaluate how the adversarial vulnerability of a model changes when successively adding more tasks (up to 5 tasks). The model is trained with all 5 tasks then we successively enable the tasks. To reduce threats to validity resulting from the choice of the task we start from and the tasks that we add, we repeat these experiments for all combinations of tasks. We show below the results for four combinations. Results for other combinations of tasks are provided in Appendix C.

%We start with a single-task model, then add successively a new task and weigh all tasks equally. We train each model and evaluate its adversarial vulnerability %against an $\mathbb{E}_{x}[\delta\mathcal{L}(\mathscr{T}',\epsilon)]$.

\begin{figure}[]

\begin{center}
\centerline{\includegraphics[width=\columnwidth]{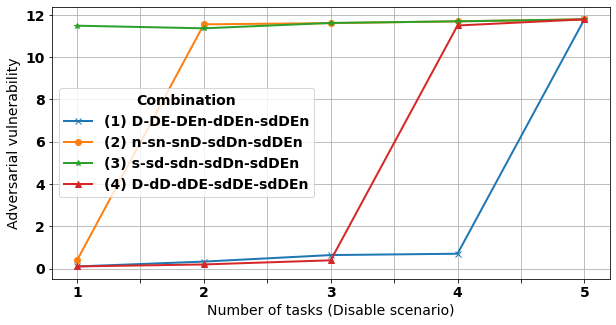}}

\caption{Adversarial vulnerability for 4 different combinations of tasks. In each combination, we enable one additional task and report the exact adversarial vulnerability of the new model. Evaluated tasks: \emph{s}: Semantic segmentation, \emph{d}: Z-depth, \emph{D}: Euclidian depth, \emph{n}: Normal estimation, \emph{E}:Edge detection. }
\label{vuln-nb-tasks}
\end{center}

\end{figure}

Figure~\ref{vuln-nb-tasks} shows the results. We observe that increasing the number of tasks often increases the vulnerability of the model. This confirms again that the main claim of \cite{mao2020multitask} %according which adding more tasks decreases model vulnerability 
does not generalize to any combinations of tasks.
%the theoretical results of Yang et al. \cite{yang2020multitask} does not generalize to any combinations of tasks. 
%This confirms again our main hypothesis that \emph{adding more tasks yields a false sense of robustness}. 
We also observe that there is no monotonic relationship between the model vulnerability and its number of tasks for cases (2) and (3), whereas in cases (1) and (4) the increase is not linear and mostly occurs at one specific point (i.e., when the segmentation task $s$ is added). More generally, across all four cases, task $s$ appears to be the main factor contributing to the increased vulnerability of the model. This supports our claim that the most marginally vulnerable tasks are the dominant factors to increasing the model vulnerability.

%marginal change of model vulnerability mostly depends on the marginal vulnerability of the added task.

%As our theorem 6 showed, adding more tasks does not change monotonically the vulnerability of the model. The vulnerability keeps decreasing for combination (2) after the first task, while it keeps increasing for combination (4), and combinations (1) and (3) show a non-monotonic relationship. It also shows that it is only when adding the segmentation task to any combination of tasks (step 1 for combination (3), step 2 for combination (2), step 4 for combination (4) and step 5 for combination (1)), that the vulnerability drastically increases, which confirms Theorem 6 insights about the relationship between the marginal vulnerability and robustness of the additional task, irrespective of the total number of tasks we attack. Finally, it shows that starting from a vulnerable task (semantic segmentation (s) in combination 2), adding robust tasks can dampen the overall vulnerability of the models.% but the impact 
%we can dampen the overall vulnerability of the model by adding robust tasks.  

\begin{mdframed}[style=MyFrame]
\textbf{
Answer to RQ2:} The marginal vulnerability increase of the %multi-task 
model mainly depends on the vulnerability of the newly added task and the most vulnerable previous task. This implies that, when tasks have equal weights, (1) the more vulnerable the tasks in the model are, the less likely adding new tasks will increase the robustness of the model; and (2) adding a vulnerable task may actually decrease %model
the robustness of the whole model.
%Our result suggest that 
%Training a model with a weak task can drastically increase the adversarial vulnerability of the whole model, indiscriminately of the number of tasks.   
%}
\end{mdframed}

\begin{table}

\begin{center}
\begin{small}
\scalebox{0.7}{
\begin{tabular}{|cc|ccccc|ccccc|}
\hline
Attack &  & \multicolumn{5}{c|}{Baseline (A)} & \multicolumn{5}{c|}{Weighted (B)}\\
\hline
     \multicolumn{2}{|c|}{Auxiliary $\rightarrow$} & s & d & D & n & E & s & d & D & n & E  \\
\hline
\multirow{5}{*}{Single} &s & \textit{0.82} & 0.86 & 0.97 & 0.96 & 0.93  & - &  0.52 &  0.57 &  0.58 &  0.54\\
& d & 5.74 & \textit{5.61} & {5.28} & 6.88 & 6.41 & 1.25 &  - &  2.00 &  1.65 &  1.92\\
& D & {5.93} & {6.14} & \textit{6.4} & 7.12 & 8.31 & 2.16 &  1.88 &  - &  2.11 &  1.78\\
& n & {7.43} & {9.48} & {8.93} & \textit{10.82} & {9.08} & 6.61 &  9.46 &  8.09 &  - &  8.14\\
& E & {12.93} & {19.29} & {18.44} & {15.16} & \textit{22.57} &  6.44 &  5.50 &  8.02 &  5.49 &  -\\
\hline
\hline
  \multirow{5}{*}{Multi} &s & - & 0.85 & 0.96 & 0.95 & 0.91 & - &  0.45 &  0.49 &  0.48 &  0.43\\
& d & {1.99} & - & {5.42} & {4.8} & {4.46} & 0.56 &  - &  2.06 &  0.82 &  0.75\\
& D & {2.14} & {6.02} & - & {5.02} & 6.07 &  0.88 &  1.74 &  - &  0.94 &  0.68\\
& n & {4.61} & {9.4} & {8.93} & - & {8.7} & 3.65 &  9.22 &  7.99 &  - &  6.32\\
& E & {7.58} & {18.5} & {17.44} & {12.48} & - & 3.24 &  4.93 &  7.00 &  3.36 &  -\\
\hline
\end{tabular}
}
\caption{Relative task vulnerability (lower is better). (A): adversarial attack against uniformly weighted tasks. (B): Adversarial attack against optimally weighted tasks. Each row refers to the main task evaluated and the column to the auxiliary task. In the top half (Single), we only attack the main task, in the bottom half (Multi), both tasks are attacked.
}
\label{table:tasks-opt}
\end{small}
\end{center}
\end{table}

\section{RQ3: Task Weight Optimization}
\label{weight_evaluation}

We evaluate how the optimization of weights of the losses of each task can be used as a defense, through the selection of optimal weights. We investigate and as an adaptive attack to overcome the gradient masking of multi-task learning.

\subsection{Robustification through optimal weights}

Given that simply adding tasks is not a promise of increased robustness, we suggest that a better way would be to adjust the weights of the tasks (in the loss function). The problem of setting the task weights has been previously studied in the context of optimizing the clean performance of multi-task models \cite{chen2018gradnorm,vandenhende2021multi,standley2020tasks}. Specifically, Zamir et al. \cite{taskonomy2018} provided the optimal weights for all combinations of tasks of their dataset (see Appendix B.). %We provide in Appendix B. the matrix of optimal weights.%we consider and shown that this increases the clean performance of any multi-task model compared to its counterpart with equal weights. 

To evaluate the potential benefits of adjusting the weights, we conduct an empirical evaluation and compare the vulnerability of models using equal task weights (the \emph{Baseline} models) with the equivalent model using the optimized task weights suggested by \cite{taskonomy2018} (the \emph{Weighted} models). More precisely, we consider any pair of tasks where the first task is the main task and the second is the auxiliary task (added for the specific purpose of making the model less vulnerable on the main task). To compare the equal-weight model with the optimized-weight model, we use the relative task vulnerability metrics as this metric is independent of the task weights. We use both single-task PGD \cite{pgd_madry2019deep} and multi-task PGD \cite{mao2020multitask}.

Results are shown in Table \ref{table:tasks-opt} (Baseline A vs Weighted B). We see that for each pair of tasks, the weighted model is less vulnerable than the baseline model. This confirms our hypothesis that a careful setting of the weights can reduce model vulnerability, even where the addition of tasks with equal weights has a negative effect. For instance, in the baseline models, we observe that the vulnerability of the model on task $s$ is lower when $s$ is the only task than when any other task is added. When weights are optimized though, the vulnerability of the weighted multi-task model on $s$ is always lower than $s$ alone \emph{regardless of the task that is added}. These results can be explained by the fact that the weight optimization proposed in \cite{taskonomy2018} aims to reduce the influence of dominant tasks during learning. As the two attacks inherently target the most dominant tasks (which have a higher average contribution to the loss function), the weight optimization improves both clean performance \emph{and} robustness.

\subsection{Adaptive Gradient Attacks}

We investigate whether weight optimization remains an effective defense against adaptive attacks. We consider Auto-PGD \cite{autoattack} -- the strongest adaptive gradient attack in the literature -- which adjust the step of the attack and the weight of the momentum at each attack iteration.

We also propose another way to make an attack adaptive. %to the task weights. 
The principle of our new attack is to weight the contribution of each task when computing the perturbation to guarantee that the attack affects all targeted tasks -- including those that have a smaller contribution to the joint loss. 

%While taking into account the dominance of tasks and weighing their losses improve the robustness of models, task-specific optimization can also be taken into account to design adaptive attack algorithms.

%Traditional gradient attacks like FGSM or PGD focus only on the gradient of one task, moving to a multi-task context requires to think about (1) which tasks to attack at each step and (2) how do we weight the contribution of each task when computing the perturbation. 

We introduce the concept of Task Attack Rate to optimally weigh the gradient of each attacked task. Task Attack Rate is inspired by the inverse task learning rate proposed by \cite{chen2018gradnorm} for the GradNorm optimization for 
%multi-task 
training. 

%While traditional gradient attacks like FGSM or PGD focus only on the gradient of one task, moving to a multi-task context requires to think about (1) which tasks to attack at each step and (2) how do we weight the contribution of each task when computing the perturbation. Taking into account these questions leads to adaptive attacks much stronger than state of the art. To design these attacks, we introduce the concept of Task Attack Rate, inspired by the inverse task learning rate proposed by \cite{chen2018gradnorm} for the GradNorm optimization for multi-task training.

\begin{definition}

%Let $\mathscr{M}$  be a multi-task model with $\mathscr{T}_{M}= \{t_1,...,t_M\}$ tasks, an input $x$, $\bar{y} = (y_1, ..., y_M)$ its corresponding ground-truth.
%We denote the set of attacked tasks $\mathscr{T}_{N}$ a subset of the model's tasks $\mathscr{T}$ and let $\mathcal{L}$ be the joint task loss of attacked tasks.

We define the \emph{Inverse task attack rate} of the task $i$ under an $\epsilon$-sized $\| .\|_p$- iterative attack on $\mathscr{T}'$ at step $t$ the loss ratio for a specific task $i$ at step $t$: 
%\begin{equation*}
    $\tilde{\mathcal{L}_i}(t) = \frac{\mathcal{L}_i(t)}{\mathcal{L}_i(0)}$
%\end{equation*}

Smaller $\tilde{\mathcal{L}_i}$ implies that task i is faster to perturb. Similarly, we define the relative inverse attack rate as
$ r_i(t) = \frac{\tilde{\mathcal{L}_i}(t)}{E_i[\tilde{\mathcal{L}_i}(t)]}$

\end{definition}

%Our attack algorithms weight the loss of each task with its associated inverse attack rate allowing us to overcome the gradient masking achieved by optimal weight optimization of the tasks. %We propose our optimization in two flavours:
We leverage this optimization in our new attack, \emph{Multi-task Weighted Gradient Attack (WGD)}, a multi-step attack where the gradient of each task is weighted by the relative inverse task attack rate. %We will show that this optimization breaks the apparent robustness of multi-task models. 
We describe full algorithm in Appendix D.
    
%\emph{Multi-task Adaptive Gradient Attack (MAGD)}, a variant of WGD where we compute the perturbation at each step of the attack on a subset of the tasks of the model. This variant is sometimes weaker than WGD, however it provides better performance than the state of the art with a fraction of the cost (30\% less with 2 tasks and 60\% less with 4 tasks).
%The intuition behind it is that multi-task models share a portion of layers, hence a portion of vulnerability. It is therefore not necessary to attack all the tasks at each iteration, especially as some tasks are strongly correlated. 

%We also evaluate Auto-PGD \cite{autoattack}, the strongest adaptive gradient attack and show that our techniques outperforms the SoTA  for most combinations.%both on attacking one and multiple tasks at once. 

%Some other ways to design adaptive attacks is to adjust the step of the attack and weight of the momentum at each iteration. Auto-PGD \cite{autoattack}, is the strongest adaptive gradient attack in the literature.% and we will show that its optimizations are complementary to our adaptive attacks; i.e.  

We empirically assess whether the two adaptive attacks can overcome the robustification mechanism based on optimal weights. Hence, we measure the relative task vulnerability of the baseline (uniformly weighted) model and the optimally  weighted model against WGD and Auto-PGD; result are shown in Table \ref{table:tasks-opt-atks}.

We observe that, on both models, WGD and Auto-PGD are much stronger than PGD (see Table \ref{table:tasks-opt}). For instance, Auto-PGD increased the error against task E up to five times in comparison with PGD on the same combination of tasks, and WGD caused up to two times more error than PGD for the combination of tasks supporting n.

Table \ref{table:tasks-opt-atks} also reveals that the weighted model is as vulnerable as the baseline model. This confirms that the adaptive attacks negate the benevolent effects of weight optimization. 

In the end, the only viable way to improve model robustness remains to add less vulnerable auxiliary tasks. Indeed, in  Table \ref{table:tasks-opt-atks} we observe that for each single-task model (i.e. the diagonal elements) there is at least one multi-task model (with the same main task) that is less vulnerable. Our previous (RQ2) conclusion remains, therefore, valid.

\begin{table}

\begin{center}
\begin{small}
\scalebox{0.65}{
\begin{tabular}{|cc|ccccc|ccccc|}
\hline
Attack &  & \multicolumn{5}{c|}{Baseline (A)} & \multicolumn{5}{c|}{Weighted (B)}\\
\hline
     \multicolumn{2}{|c|}{Auxiliary $\rightarrow$} & s & d & D & n & E & s & d & D & n & E  \\
\hline
\multirow{5}{*}{APGD} &s & 0.89  &  0.91  &  0.90  &  0.88  & 0.92  & 0.89 &  0.92 &  0.93 &  0.88 &  0.90\\
& d &  17.17 &  23.88 &  13.50 &  24.10 & 24.98 & 18.27 &  23.19 &  13.08 &  15.92 &  23.9\\
& D &  15.50 &  15.08  & 20.15 &  26.00  & 22.74 & 20.72 &  24.93 & 18.29 &  28.21 &  23.68\\
& n & 12.99 &  17.76 &  17.27 &  19.02 & 17.24 & 12.35 &  17.14 &  16.72 &  18.49 &  16.46\\
& E & 135.4 & 171.8 & 159.7 & 138.8 & 81.77 & 125.8 & 78.65 & 377.8 & 110.4 & 68.06\\
\hline
\hline
  \multirow{5}{*}{WGD} &s &  0.90 &  0.91 &  0.91 &  0.90 &  0.91 & 0.90 & 0.93 & 0.94 & 0.93 &  0.91\\
& d & 12.86 & 13.57 & 12.39 & 16.18 & 18.13 & 12.66 & 13.55 & 12.8 & 11.7 & 12.96\\
& D & 14.05 & 14.04 & 15.57 & 17.03 & 19.24  & 15.65 & 14.85 & 15.55 & 16.95 & 13.56\\
& n & 13.05 & 17.06 & 16.32 & 18.12 & 16.57 & 17.00 & 20.26 & 17.32 & 18.13 &  17.68 \\
& E & 45.35 & 90.67 & 86.04 & 57.43 & 90.19 & 106.9 & 89.89 & 116.3 & 71.42 & 90.59 \\
\hline
\end{tabular}
}
\caption{Relative task vulnerability under two different attacks (lower is more robust). (A): adversarial attack against uniformly weighted tasks. (B): Adversarial attack against optimally weighted tasks. Each row refers to the task attacked and evaluated and the column the auxiliary task.
}
\label{table:tasks-opt-atks}
\end{small}
\end{center}
\end{table}

\begin{mdframed}[style=MyFrame]
{
Answer to RQ3:} %Adjusting the task weights is a key lever for improving the model robustness. However, adaptive attacks that take into account the weight balancing can overcome this gradient masking and confirms that multi-task learning provides a limited defense against adversarial examples.
Weight optimization in multi-task learning decreases model vulnerability against non-adaptive attacks only. Therefore, the only way to improve the robustness of multi-task models remain to add less vulnerable auxiliary tasks.%Indeed, adaptive attacks that take into account the weight balancing can overcome this gradient masking defense and confirms that overall multi-task learning is a weak defense against adversarial examples. 

%Fortunately, existing task weight optimization methods targeting clean %model performance appears effective when considering robustness.
%}
\end{mdframed}
\section{RQ4: Task Selection}

Our previous results imply that one should carefully select the auxiliary tasks added to reduce model vulnerability. Generally speaking, the addition of auxiliary tasks can even have negative effects. Auxiliary task selection, however, comes with three drawbacks \cite{standley2020tasks}: the size of the model is bigger (due to the addition of the task-specific decoder), the convergence of the common encoder layers is slower, and the clean performance risk deteriorating as more tasks are added. This raises the question of how to select the combination that yields the lowest vulnerability \emph{without} evaluating the vulnerability of all the possible combinations. 

We propose three methods to make this selection more efficient. Their common idea is to compute a proxy of the adversarial vulnerability which is fast to get and correlated to the real adversarial vulnerability. As before, we consider each possible pair of tasks with one main task and one auxiliary task. We use the relative task vulnerability of the models on the main task, as this metric is independent of the task (unlike, the task loss whose scale depends on the task). 

We use the Pearson coefficient to measure correlation as it captures the linear relationships between variables. We compute the correlation between the relative task vulnerability of one combination of tasks in the expensive model with the relative task vulnerability of the same combination on the cheaper surrogate.

The first method is \emph{early stopping}, that is, training the model after a predefined (small) number of epochs. Here, we stop after 50 epochs while the full training lasts for 150 epochs. Strong correlations between the vulnerability of the models trained over these two number of epochs would indicate that one can decide which task combination is optimal after few epochs.

The second is to use a surrogate (less expensive) encoder and evaluate all task combinations on this encoder. We hypothesize that combinations of tasks that are effective when joint to the surrogate encoder are also effective with the original model. We validate this hypothesis by measuring the correlation between the vulnerability of the models formed by joining the surrogate encoder with any combination of tasks and the vulnerability of their counterparts using the original encoder. We use ResNet18 as the surrogate encoder and ResNet50 as the target encoder.

Our third selection method is guided by the clean performance that the model achieves on the main task when the auxiliary tasks are added. The existence of a correlation between clean performance and vulnerability would allow avoiding the cost of evaluating adversarial vulnerability (e.g. applying PGD) and reuse existing results on clean performance predictions \cite{standley2020tasks} to predict adversarial vulnerability. 

Table \ref{tbl:stats-atks} shows the Pearson correlation coefficient with the associated p-value. We observe that all three proxy methods are correlated with the real adversarial vulnerability values. Specifically, early stopping offers a medium correlation (0.55) while the methods based on the surrogate encoder and the clean performance achieve a very strong correlation (0.94).

\begin{mdframed}[style=MyFrame]
\textbf{
Answer to RQ4:} While exhaustively computing the adversarial vulnerability for all task combinations is computationally expensive, this cost can be significantly reduced while obtaining a good approximation of the model vulnerability. In particular, guiding the auxiliary task selections by the clean performance or the vulnerability of a smaller surrogate model offers reliable indications of the benefits achieved by adding these tasks.
\end{mdframed}

\begin{table}[]
\begin{center}
\begin{small}

\begin{tabular}{|c|c|c|c|}
\hline
  Proxy & Target & Pearson & p-value  \\
  \hline
  {50 epochs} &  {150 epochs} & 0.55 & 4.23e-3\\
  {Resnet18} & {Resnet50} & 0.94 & 1.42e-12\\
  {Clean performance} & {Robustness} & 0.98 & 1.91e-17\\
  %{MT Robustness} &  {Clean performance} & 0.94 & 2.58e-12\\
  %{ST Robustness} &  {MT Robustness} & 0.93 & 8.41e-12\\
   
\hline
\end{tabular}

\end{small}
\end{center}

\caption{Pearson correlation between the real adversarial vulnerabilities and proxy values from three different methods.}
\label{tbl:stats-atks}

\end{table}
%\input{LaTeX/-dense}

%------------------------------------------------------------------------
%\input{LaTeX/-classification}

%------------------------------------------------------------------------
%\section{Threats to validity}

%Our study focused only on one dataset to evaluate the factors behind the robustness of multi-task models, however, we proposed a theoretical evaluation that relies on the least number of hypotheses to cover all multi-task learning paradigm. We also studied a large number of tasks and a diverse set of architectures, with different families and different sizes.

%To mitigate the risk of coding errors, we base our implementation and experiments on existing software. The training of the models rely on the models and architectures shared by \cite{taskonomy2018}. The source code to evaluate the adversarial vulnerability is the one proposed by \cite{simon2019first}. 
%We provide all our algorithms, models, and open source-code at \emph{[Anonymized for double-blind review]} 

%------------------------------------------------------------------------
\section*{Conclusion}

We have presented what is to date the largest evaluation of the vulnerability of multi-task models to adversarial attacks.  
Our study does not entirely reject the benefits of adding auxiliary tasks to improve robustness, but rather tones down the generality of this proposition. We show that what matters is the inherent vulnerability of the tasks which, in turn, implies that robustness can be improved by carefully selecting the auxiliary tasks to add and adjusting the task weights. 

%We also reveal that while multi-task models can be robust to some norms and few steps attacks, they are inherently weak against multi-steps and $l_\infty$ attacks. We propose two adaptive attacks that are stronger and more efficient than state-of-the art and show they can overcome the gradient masking defense granted by multi-task learning. 

We evaluate different settings of multi-task learning, with a large combination of tasks, architectures, attack strengths and norms and show that in multiple settings, multi-task learning fails to protect against gradient attacks. 

We also demonstrate that weight optimization can significantly improve the robustness of multi-task models, however, falls short to protecting against adaptive attacks for some tasks. In particular, we propose a new adaptive attack, WGD, that balances the gradient of the tasks and overcomes the gradient masking defense of multi-task learning.    

Taking the perspective of the defender, we show that one can identify the most robust combinations of tasks efficiently by working on cheap surrogates (i.e. fewer training epochs or smaller architectures).% and that the fine-tuning of the weights of the tasks yields an improved robustness.% and that the combinations that are best suited for transfer learning are not the ones that help to robustify the models.

%On the attacker side, we provided cheap optimizations that easily overcome the improved robustness of multi-task models while reducing the cost of the attacks. We also hint that while multi-task models can be robust to some norms and few steps attacks, they are inherently weak against multi-steps and $L-\infty$ attacks.

Overall, our research contributes to guiding practitioners in the development of robust multi-task models and paves the way for methods to improve together the clean performance and the robustness of multi-task models. 

%------------------------------------------------------------------------

% Use \bibliography{yourbibfile} instead or the References section will not appear in your paper
\bibliography{egbib}

\end{document}